\documentclass[runningheads,orivec,a4paper]{llncs}
\usepackage{graphicx}
\usepackage{hyperref}
\usepackage{amsmath}
\newcommand{\figref}[1]{Fig.~\ref{#1}}
\newcommand{\Figref}[1]{Figure~\ref{#1}}
\newcommand{\tabref}[1]{Table~\ref{#1}}

\begin{document}

\title{Permutohedral Attention Module \texorpdfstring{\newline}{} for Efficient Non-Local Neural Networks}
\titlerunning{PAM for Efficient Non-Local Neural Networks}
%
\author{Samuel Joutard, Reuben Dorent, Amanda Isaac, Sebastien Ourselin,\texorpdfstring{\newline}{} Tom Vercauteren, Marc Modat}
%
\authorrunning{S. Joutard et al.}
%
\institute{School of Biomedical Engineering \& Imaging Sciences, King's College London}
%
\maketitle              
\begin{abstract}
Medical image processing tasks such as segmentation often require capturing non-local information. As organs, bones, and tissues share common characteristics such as intensity, shape, and texture, the contextual information plays a critical role in correctly labeling them. Segmentation and labeling is now typically done with convolutional neural networks (CNNs) but the context of the CNN is limited by the receptive field which itself is limited by memory requirements and other properties. In this paper, we propose a new attention module, that we call Permutohedral Attention Module (PAM), to efficiently capture non-local characteristics of the image. The proposed method is both memory and computationally efficient. We provide a GPU implementation of this module suitable for 3D medical imaging problems. We demonstrate the efficiency and scalability of our module with the challenging task of vertebrae segmentation and labeling where context plays a crucial role because of the very similar appearance of different vertebrae. 

\end{abstract}
\section{Introduction}
Convolutional neural networks (CNNs) have become one of the most effective tools for many medical image processing tasks such as segmentation. However, working with medical images has its own idiosyncratic challenges. The organs, tissues or bones can have very similar characteristics, such as intensity, texture, or shape. As a consequence, the differentiating aspects of each individual structure come from the context and the position of the item of interest in the larger surroundings. However, naively extracting non-local characteristics of a region requires much more computation and memory than focusing on its local characteristics.
This currently makes using non-local context highly non-trivial in medical imaging.
Hence, an efficient approach to exploit non-local characteristics in deep learning could transform several medical imaging pipelines. 

The notion of contextual information is intimately related to the concept of receptive field in deep learning. The receptive field of an output variable corresponds to the region in the input influencing its value. Recent studies on receptive field in CNNs~\cite{study_on_receptive_field} have proven that the receptive field size is sub-linear in the number of convolutional layers. In order to improve the receptive fields of a CNN, two main solutions have been adopted: down-sampling layers and dilated convolutions~\cite{dilated}. Use of down-sampling layers efficiently increases the receptive field size but decreases the resolution of the information. Hence, it is not suitable for very granular segmentation in which case dilated convolutions are often preferred~\cite{Wenqi}. Both of these solutions result in a fixed receptive field, 
which means that all contextual information in the receptive field will be taken into account whether it is relevant or not. Attention modules have been used to prune irrelevant information in medical imaging~\cite{attention_gate,weak_attention_2}. 
Yet, these tools remain suboptimal as they do not allow to capture large scale context.
However, the extended self-attention formulation of~\cite{non_local_neural_networks} offers a solution to dynamically adapt the individual receptive field of each output variable to only make use of relevant non-local information. Despite its attractive properties, this formulation of self-attention has not yet been applied to medical images partly because its computational requirements scale as $O(N^2)$ ($N$ is the number of voxels). 

In this paper, we propose a new self-attention module called Permutohedral Attention Module (PAM), which makes use of the efficient approximation algorithm of the Permutohedral Lattice~\cite{PL}. We adapted the algorithm of~\cite{PL}, originally designed to perform denoising, into a trainable self-attention module able to capture and process contextual information. The Permutohedral Lattice algorithm was previously used in a trainable framework in the more general context of sparse high dimensional convolutions for computer vision \cite{SparseBilConv}. The self-attention approach is a suitable compromise for medical image processing in terms of memory and computation between \cite{SparseBilConv} and standard convolutions while preserving most of the model representation capacity increase of \cite{SparseBilConv} to process contextual information. Our module, similarly to the original non-local self-attention mechanism formulation, dynamically adapts the receptive field of each output variable in a learned way while being, in contrast to~\cite{non_local_neural_networks}, applicable to medical images as it has low memory requirements, computationally scaling as $O(N)$. We evaluate our module on the challenging task of vertebrae segmentation. Vertebrae segmentation aims to label each individual vertebra and is used in practice as an initial step of various pipelines such as modality fusion, spine surgery planning and surgical guidance. As consecutive vertebrae have very similar local appearance, non-local information is compelling to identify them. 

In Section 2, we first define self-attention and how it has been used, we then introduce the PAM. In section 3 we first highlight the capability of our module to capture and process contextual information without requiring a deep architecture. Then, we demonstrate its capability to improve state of the art segmentation architectures for vertebrae segmentation.

\section{Methods}

\subsubsection{The self-attention mechanism}
Self-attention used in deep learning frameworks can be defined as follows: consider a standard deep learning framework where the input $\mathbf{x}=(x_1,..x_N)$ is processed first by a section of the network we call descriptor network $\mathbf{v}_{\psi}$, and then by the rest of the network we call prediction network $g_\theta$ ($\psi$ and $\theta$ are the respective parameter sets). The model predicts $\mathbf{y}$ so that:
\begin{equation}
\mathbf{y}=g_{\theta}(\mathbf{v}_{\psi}(\mathbf{x}))
\end{equation}
We define $A_{\phi}(.)$ the self-attention mechanism parameterized by $\phi$ which combines the non-local input descriptors in a learned way. For all input $\mathbf{x}$, $A_{\phi}(\mathbf{x})$ is a $N\times N$ self-attention matrix where the coefficient $A_{\phi}(\mathbf{x})_{i,j}$ characterizes the attention of $x_i$ towards $x_j$. Our framework including an attention mechanism predicts $\mathbf{y}_{att}$:
\begin{equation}\label{eq:y_att}
    \mathbf{y}_{att}=g_{\theta}(A_{\phi}(\mathbf{x})\cdot \mathbf{v}_{\psi}(\mathbf{x}))
\end{equation}
where $\cdot$~represents the matrix multiplication operator.
This formulation has two principal strengths; it can increase the receptive field of each output variable up to the whole input, and it can modulate the receptive field of each output variable with respect to the input characteristics. 
To our knowledge, attention modules in deep learning either compute the entire self-attention matrix on a low dimensional input or use a local attention mechanism that can be seen as a strong approximation of the non-local self-attention formulation.
Specifically in the medical imaging context, previous works~\cite{attention_gate,weak_attention_2,roy2019recalibrating} implicitly used a simplification of \eqref{eq:y_att} with a diagonal self-attention matrix. This solution can be applied to large images since it scales linearly with the number of voxels but does not help to capture contextual information. 

Different implementations of the non-local self-attention matrix are listed in~\cite{non_local_neural_networks}. These can be unified as follows:
\begin{equation}
    A_{\phi}(\mathbf{x})=[\gamma(\phi_1(x_i)^T \cdot \phi_2(x_j))]_{1 \leq i,j \leq N}
\label{standard attention}
\end{equation}
where $\phi=(\phi_1,\phi_2)$ is a pair of embedding functions (possibly identities) and $\gamma$ is typically either identity, exponential or ReLU. Hence, these approaches are impractical to apply to 3D images because the number of interactions to be computed scales as $O(N^2)$.

\subsubsection{Permutohedral Attention Module}
The proposed PAM relies on a slightly different formulation of the self-attention matrix to align more closely with the formulation of the non-local means filtering algorithm~\cite{non_local_means} used in the denoising literature. When applied to the set of feature-descriptor pairs $(f_i, v_i)_{i\leq n}$ (where $n$ is the number of variables described), non-local mean gives the set of filtered descriptors:
\begin{equation}\label{eq:nlm}
    \forall 1 \leq i\leq n, \quad v'_i = \sum\limits_{j=1}^n \exp(-|| f_i -f_j||_2) v_j
\end{equation}
Hence:
\begin{equation}\label{eq:attentionmatrixform}
A_{\phi}(\mathbf{x})=[\exp(-||f_{\phi}(x_i) -f_{\phi}(x_j)||_2)]_{i,j\leq N}
\end{equation}
is the corresponding attention formulation with $f_{\phi}$ a feature extractor network ($\phi$ its parameter set). 

\begin{figure}[tb!]
\centering
\includegraphics[width=0.9\textwidth]{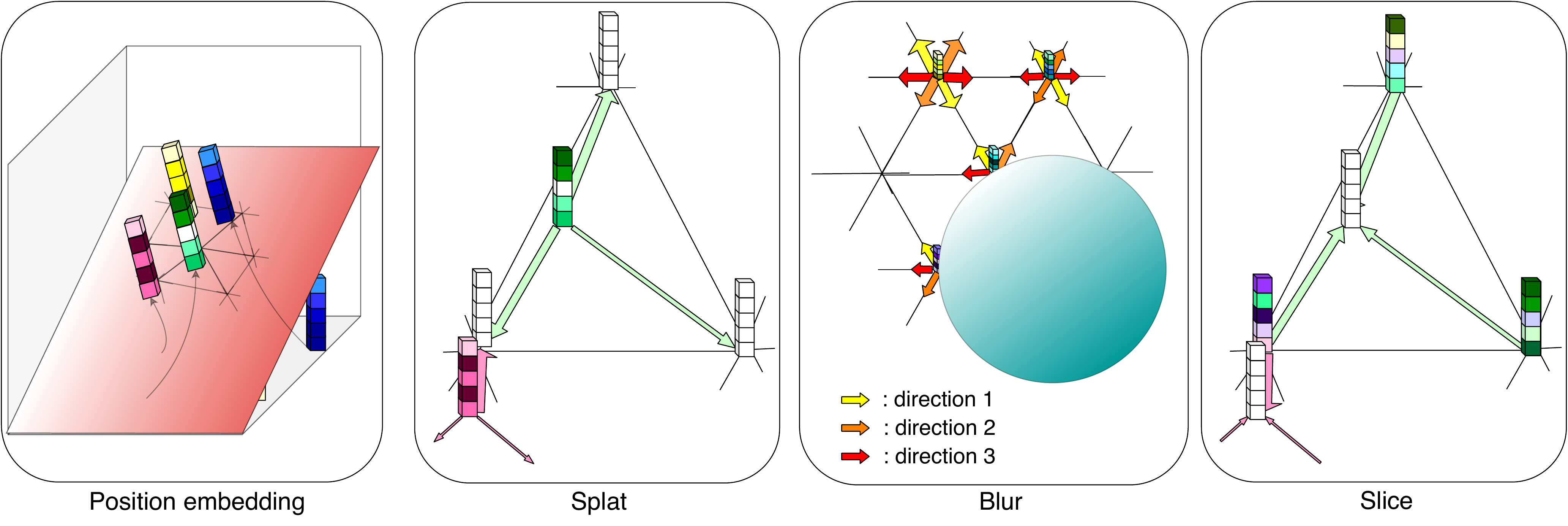}
\caption{The features lying in $\rm I\!R^f$ are embedded in a hyperplane of $\rm I\!R^{f+1}$ to position each variable. This hyperplane is partitioned in simplices by a mesh called the Permutohedral Lattice. The Splat phase describes the vertices of the Permutohedral Lattice based on the neighbouring variables. The Blur step applies a Gaussian blur along each direction consecutively. Finally, the Slice step re-projects the filtered descriptors from the vertices to the variables.} \label{PL}
\end{figure}

Avoiding a brute-force computation of \eqref{eq:nlm}, we adapted the Permutohedral Lattice approximation algorithm~\cite{PL} to estimate the self-attention module output $\mathbf{v'}_{\phi, \psi}(\mathbf{x})=A_{\phi}(\mathbf{x})\cdot \mathbf{v}_\psi(\mathbf{x})$ in $O(N)$ against $O(N^2)$ for the original non-local neural network formulations listed in~\cite{non_local_neural_networks}. Learning the parameter sets $\phi$ and $\psi$ is achieved through back-propagation. Hence, the PAM can be integrated in a deep learning framework to compute self-attention for high dimensional inputs (cf. Section~\ref{sec:models} for concrete architectures examples). The PAM approximates the proposed attention mechanism in 4 steps: embedding of the features into the Permutohedral Lattice higher dimensional space, Splat, Blur and Slice, as illustrated in \figref{PL}. Each of these steps scales linearly in $N$.

The advantage of this approximation algorithm against other possibilities~\cite{GKDtree,gride_filtering} is that the gradients with respect to the input feature vectors $\mathbf{f_{\phi}(\mathbf{x})}$ and the descriptor vectors $\mathbf{v_{\psi}(\mathbf{x})}$ can be expressed using the four steps composing the forward pass and be fully parallelized. Omitting the dependencies in $\mathbf{x}$, $\phi$, and $\psi$, we can express the forward pass as:
\begin{equation}
    \mathbf{v'} = \; \mathcal{S}l(\mathcal{B}(\mathcal{S}(\mathbf{\mathcal{E}(f)}, \mathbf{v})), \; \mathbf{\mathcal{E}(f)})
\end{equation}
where $\mathcal{E}$ is the embedding operator, $\mathcal{S}$ is the Splat operator, $\mathcal{B}$ is the Blur operator and $\mathcal{S}l$ is the Slice operator. With the same notations, the backward pass can be expressed as:
\begin{align}
    \frac{\partial L}{\partial \mathbf{v}} &=\frac{\partial L}{\partial \mathbf{v'}} \circ \frac{\partial \mathbf{v'}}{\partial \mathbf{v}} = \; 
        \mathcal{S}l(\tilde{\mathcal{B}}(\mathcal{S}(\mathbf{\mathcal{E}(f)}, \frac{\partial L}{\partial \mathbf{v'}})), \; \mathbf{\mathcal{E}(f)}) 
\label{small equation}
\\
%
\frac{\partial L}{\partial f_i} = \frac{\partial L}{\partial \mathbf{v'}} \circ \frac{\partial \mathbf{v'}}{\partial f_i}  &=
E^T\cdot
\left[\tilde{\mathcal{B}}\mathcal{S}\mathcal{E}(\mathbf{f}, \frac{\partial L}{\partial \mathbf{v'}})_{\sigma_i^j}\cdot v_i^T + \mathcal{B}\mathcal{S}\mathcal{E}(\mathbf{f}, \mathbf{v})_{\sigma_i^j} \cdot \left ( \frac{\partial L}{\partial \mathbf{v'}_i}\right )^T\right]_{j\leq(f+1)}
\label{big equation}
\end{align}
where $L$ is the loss and $\forall a,b \; \mathcal{B}\mathcal{S}\mathcal{E}(a,b)=\mathcal{B}(\mathcal{S}(\mathcal{E}(a),b))$ (similarly with $\mathcal{\tilde{B}}\mathcal{S}\mathcal{E}$). $\tilde{\mathcal{B}}$ is the Gaussian blurring operator where the Gaussian blur is applied in the reverse order in terms of direction of the Lattice. $E$ is the position embedding matrix and $\sigma_i$ is a permutation computed during $\mathcal{E}$.

\section{Experiments}
\subsubsection{Data}
We evaluate the impact of PAM for non-local neural networks for the task of simultaneous segmentation and labeling of vertebrae.
We performed our experiment on the CSI 2014 workshop challenge data\footnote{\url{http://spineweb.digitalimaginggroup.ca/}}, which consists of 20 CT images. We used all 20 CT images in our framework using a 5-fold cross validation for evaluation. We resampled the data to obtain (1mm, 1mm, 3mm) voxels. 

\subsubsection{Implementation details}
We implemented the PAM as well as all our pipelines using Pytorch. We optimized our networks with ADAM on $160\times160\times96$ patches with a fixed learning rate of $0.001$ and a batch size of 1. We used the Dice loss as loss function. Our implementation is publicly available\footnote{\url{https://github.com/SamuelJoutard/Permutohedral_attention_module}}.

\subsubsection{Models}\label{sec:models}
As a preliminary experiment, we consider a specific 6-layer fully convolutional network~(referred to as FCN). We design 2 baselines for this shallow setting. FCN is a plain fully convolutional network with a first ($3\times3\times3$) convolution with 18 output channels followed by 4 ($3\times3\times3$) embedding convolutions with 18 output channels each and a prediction ($1\times1\times1$) convolutional layer. Dil.FCN is a similar architecture where we replace each embedding convolution by a dilated block. A dilated block corresponds to 3 ($3\times3\times3$) convolutions in parallel with 6 output channels each. Of these 3 convolutions, two have dilated filters (dilatation factor of 2 and 4 respectively). The outputs of a dilated block are then concatenated before the next block. Then, we incorporate in each baseline the PAM (networks are respectively called FCN+PAM and Dil.FCN+PAM) and compare the results of those 4 configurations. 

\begin{figure}[htb!]
\centering
\includegraphics[width=.9\textwidth]{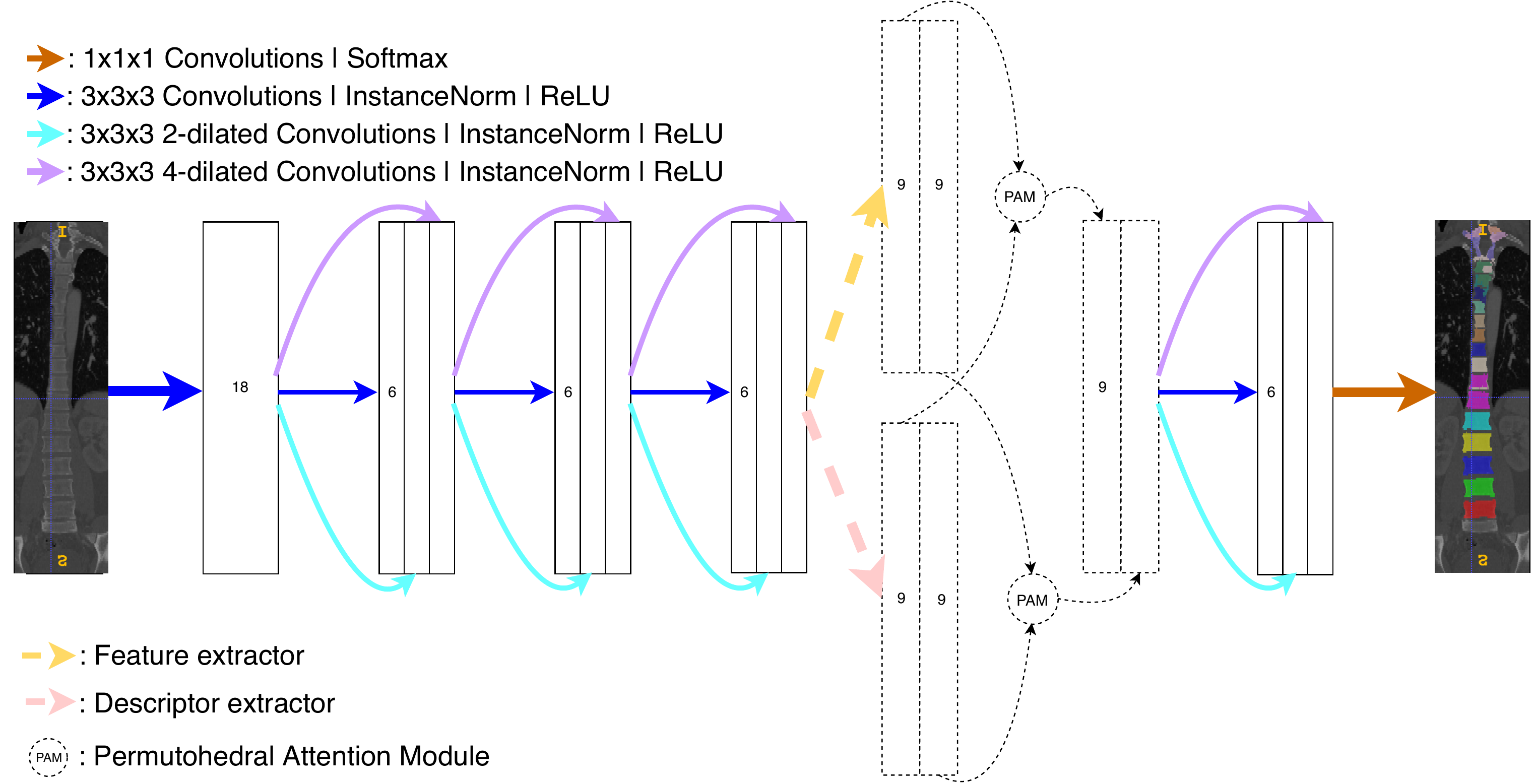}
\caption{Dil.FCN+PAM, a shallow architecture including dilated convolutions and PAM. The feature extractor and descriptor extractor are ($1\times1\times1$) convolutions. The feature extractor incorporates a mesh of spatial coordinates before applying its convolution, and is followed by a Leaky-ReLU activation function. The number on intermediate results correspond to the number of channels. We call the combination of the dashed elements Permutohedral block.} \label{SimpleNet}
\end{figure}

\Figref{SimpleNet} represents the Dil.FCN+PAM architecture. In this figure, we observe that, once we obtain the features $\mathbf{f}=(f_1, ..f_N)$ and descriptors $\mathbf{v}=(v_1, ..v_N)$ to compute attention, we split each feature and descriptor vector in two. Hence we obtain two sets of feature-descriptor pairs ${(f_i^0,v_i^0)}_{i \leq N}$ and ${(f_i^1,v_i^1)}_{i \leq N}$ on which we apply the PAM independently. There are two main advantages to doing so. First, it allows us to further reduce computation time and memory footprint. Second, it generates a per-group-of-channel attention map which makes the model more flexible (as a unique attention matrix for all descriptor channels is a particular case of two attention matrix, one for each group of channel). The reason for not splitting the feature-descriptor pairs set into more subsets is because we want a trade-off between the advantages described above and the preservation of relevant features to compute attention.

\begin{figure}[htb!]
\centering
\includegraphics[width=0.90\textwidth]{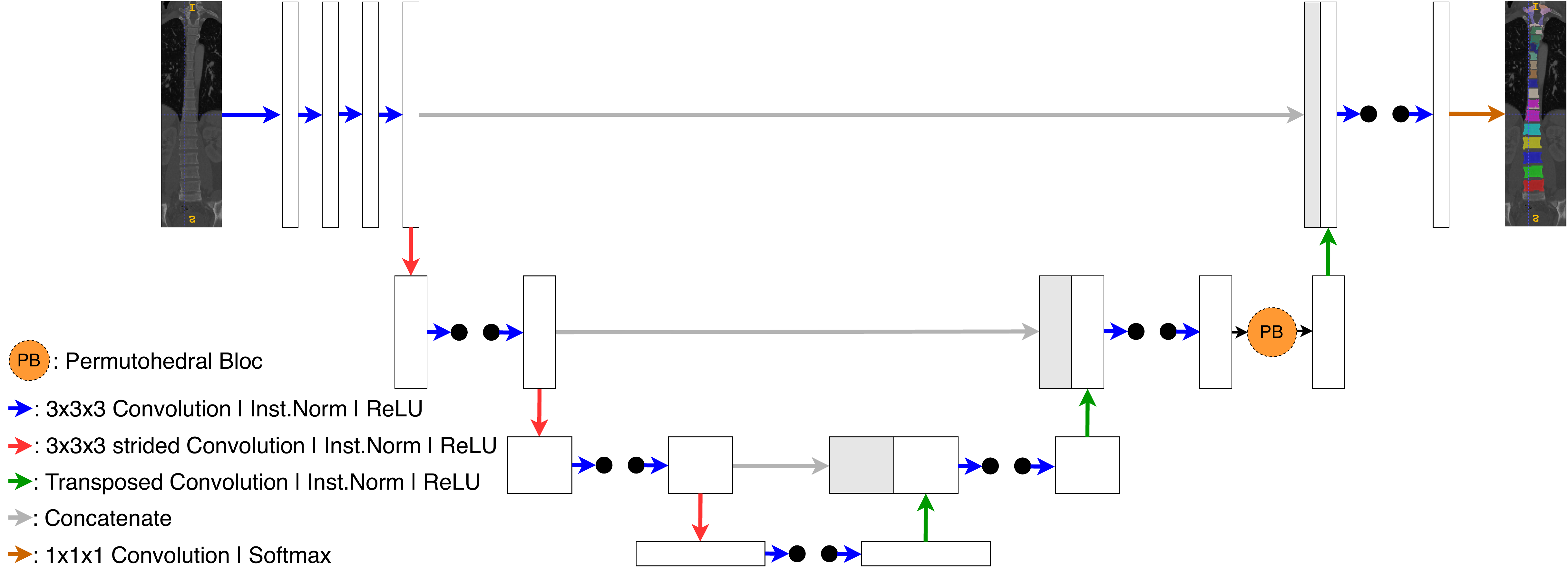}
\caption{U-PAM-Net. We make use of the Permutohedral block defined in \figref{SimpleNet}.} 
\label{unet}
\end{figure}

Then, we consider a 3D U-Net~\cite{3D_Unet} which is one of the most popular architectures for segmentation~\cite{nnunet}. We refer to our 3D U-Net simply as U-Net. We incorporate the PAM into our U-Net as shown in \figref{unet} and demonstrate that the PAM can also improve architectures which have large receptive fields (we call this network U-PAM-Net). As shown in \figref{unet}, we incorporate the PAM at the half-resolution level. Hence, we compute attention for ($2\times2\times2$) voxel regions which, in our experiments, led to similar results as computing attention at the voxel level while decreasing computation time and making convergence faster.

As the PAM introduces a small number of extra parameters, we compensate with additional channels in the first convolution on the architectures without the PAM so that the corresponding networks have either as many as or more degrees of freedom than networks with the PAM integrated.

\subsubsection{Results}
\begin{figure}[t!]
\centering
\includegraphics[width=0.9\textwidth]{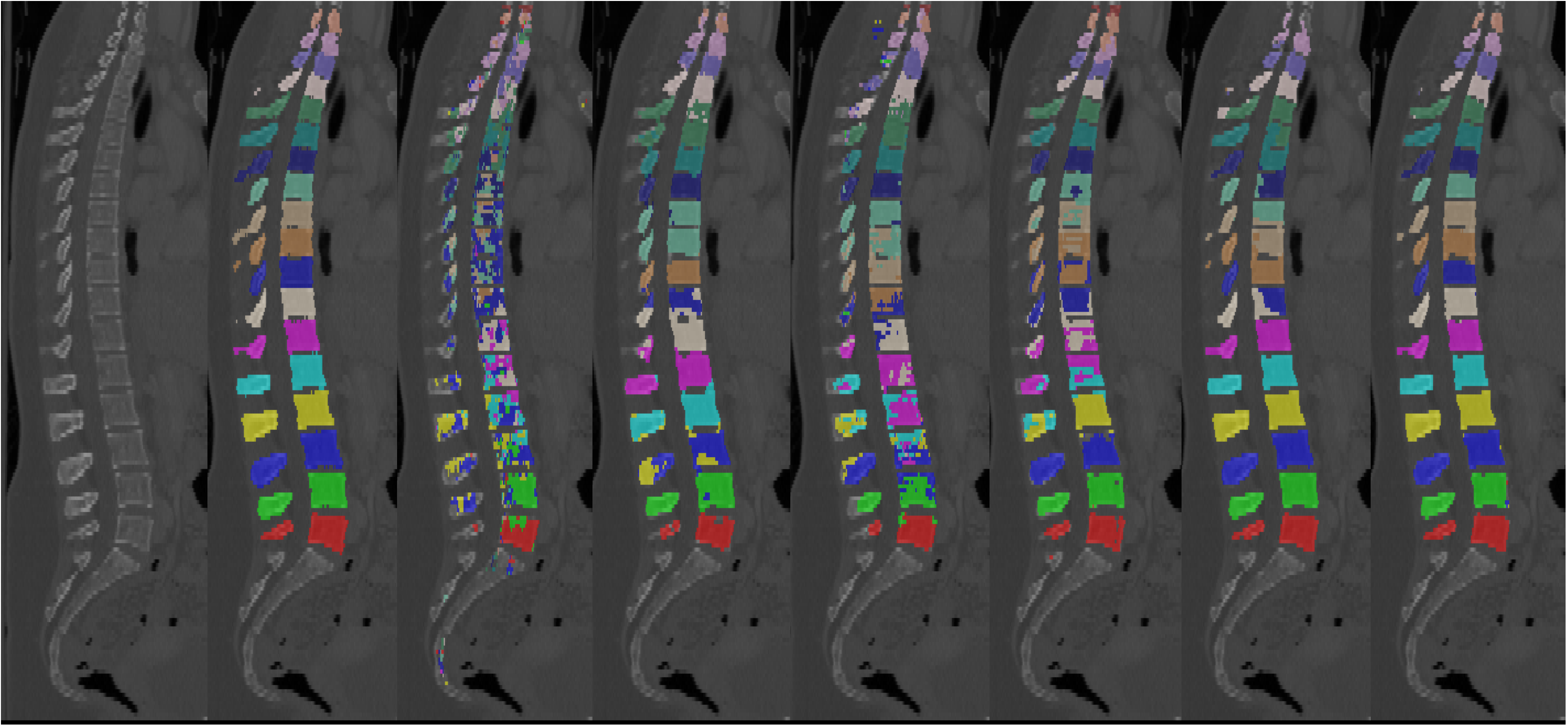}
\caption{Example of segmentation obtained by our different networks. In the corresponding order: Input slice, ground truth, FCN, FCN+PAM, Dil.FCN, Dil.FCN+PAM, U-Net, U-PAM-Net.} 
\label{comparison}
\end{figure}

We measure the performance of the different architectures with the Dice scores. \tabref{results} shows that the PAM improves performance for all the architectures it was incorporated into. In addition, we highlight that the shallow network Dil.FCN+PAM performs almost as well as the much deeper network 3D U-Net. Indeed, the dilated convolutions manage to describe the voxels using contextual information while the PAM uses those meaningful features to compute voxels interactions. \tabref{results} also illustrates the limitation of down-sampling layers pointed earlier as U-Net performs poorly on cervical vertebrae which appear very small in our images. U-PAM-Net manages to reach higher accuracy performances than~\cite{spine_hard_tuned}, which makes use of a task-specific framework especially tuned to "count" the vertebrae from spine segmentation. While ~\cite{spine_hard_tuned} report an accuracy of 81\%, our proposed framework obtained 89\% using the same evaluation metric and on the same dataset. It should be noted that the training frameworks in terms of test-train split were different for both approaches. 
\Figref{comparison} shows a representative example of the results we observed. 

\begin{table}[t!]
\begin{center}
\caption{Mean(std) Dice score (\%) of the different networks tested}
\resizebox{\textwidth}{!}{
\begin{tabular}{|c||c|c||c|c||c|c|}
\hline
Network &  FCN & \textbf{FCN+PAM} & Dil.FCN & \textbf{Dil.FCN+PAM} & U-Net & \textbf{U-PAM-Net}\\
\hline
\hline
Full & $28 (3)$ & $\mathbf{54}(14)$ & $49(7)$ & $\mathbf{70}(9)$ & $72  (9)$ & $\mathbf{81}  (9)$ \\
\hline
Cervical & $45(30)$ & $\mathbf{53}(37)$ & $\mathbf{60}(27)$ & $56(33)$ & $25(38)$ & $\mathbf{55}(39)$\\
Thoracic & $21(3)$ & $\mathbf{46}(14)$ & $42(10)$ & $\mathbf{67}(12)$ & $67(13)$ & $\mathbf{80}(12)$\\
Lumbar & $29(4)$ & $\mathbf{69}(19)$ & $58(8)$  & $\mathbf{79}(12)$ & $\mathbf{93}(2)$ & $91(5)$\\
\hline
\end{tabular}}
\label{results}
\end{center}
\end{table}

\section{Discussion}

In this work, we propose the Permutohedral Attention Module, a computationally efficient attention module to be applied in 3D deep learning framework. The PAM can be incorporated in any CNN architecture. We demonstrated its ability to efficiently handle non-local information in the context of vertebrae segmentation and presented its potential to reduce networks size in specific tasks.
Future work will notably include the investigation of asymmetric attention matrix for feature filtering and the integration of the PAM formulation in path training.

\subsubsection{Acknowledgement}
We thank E. Molteni, C. Sudre, B. Murray, K. Georgiadis, Z. Eaton-Rosen, M. Ebner for their useful comments. This work is supported by the Wellcome/EPSRC Centre for Medical Engineering [WT 203148/Z/16/Z]. TV is supported by a Medtronic / RAEng Research Chair [RCSRF1819/7/34].

%
%
%
\bibliographystyle{splncs04}
\bibliography{sam_bib_miccai}

\end{document}